# Identifying Similarities in Epileptic Patients for Drug Resistance Prediction

David Von Dollen

**Abstract**

*Currently, approximately 30% of epileptic patients treated with antiepileptic drugs (AEDs) remain resistant to treatment (known as refractory patients). This project seeks to understand the underlying similarities in refractory patients vs. other epileptic patients, identify features contributing to drug resistance across underlying phenotypes for refractory patients, and develop predictive models for drug resistance in epileptic patients.*

*In this study, epileptic patient data was examined to attempt to observe discernable similarities or differences in refractory patients (case) and other non-refractory patients (control) to map underlying mechanisms in causality. For the first part of the study, unsupervised algorithms such as Kmeans, Spectral Clustering, and Gaussian Mixture Models were used to examine patient features projected into a lower dimensional space. Results from this study showed a high degree of non-linearity in the underlying feature space. For the second part of this study, classification algorithms such as Logistic Regression, Gradient Boosted Decision Trees, and SVMs, were tested on the reduced-dimensionality features, with accuracy results of 0.83(+/-0.3) testing using 7 fold cross validation. Observations of test results indicate using a radial basis function kernel PCA to reduce features ingested by a Gradient Boosted Decision Tree Ensemble lead to gains in improved accuracy in mapping a binary decision to highly non-linear features collected from epileptic patients*

**Introduction and Motivation**

As rapid advances are made in the field of machine learning and big data, researchers are turning to these tools to develop a means to predict outcomes for patient treatment. Within the area of epilepsy research, there is much interest in understanding the underlying causes for failures in surgical treatments, as well as failures in antiepileptic drug treatments for patients.

Specific interest in finding the right treatment to combat intractable epilepsy remains at the forefront of current research. Intractable epilepsy, where there is little to no control over seizures using medications, can have a major effect on a patient's quality of life. Patients may have trouble with work or school, and may obtain injuries as the result of unpredictable onsets. Additionally, Pardoe HR et al. showed that intractable epilepsy might advance brain aging by up to 9 years [4]. Loescher et.al describe various hypotheses for causality for drug resistance, in particular the gene variant hypothesis, where it has been shown the P-glycoprotein, which is regulated by the ABCB1 gene, may play a role in drug resistance [3].

In regards to applying machine learning techniques for classification and causality discovery, the evidence is sparse in the literature. Hernández-Ronquillo et.al showed that some features of refractory patients might have predictive quality when used for a logistic regression analysis [6].

**Problem Formulation**

For this study, two hypotheses were developed:

1. *Clustering techniques and/or graph analysis and algorithms may be used to effectively model the distinct segments in refractory patients and other epileptic patients in remission. Underlying phenotypes in refractory patients may be identified and clustered.*

2. *Significant classification accuracy for patients likely to develop drug resistance may be established and refined by predictive modeling.*

Metrics for evaluation for classifiers included AUC and Cross-Validation Accuracy. Metrics for clustering included adjusted rand score and adjusted mutual information.

For training, *K*-fold Cross-Validation was used to partition training samples into *k* folds, and retained *k*-1 partitions as training data. This process was repeated for *k* times, with each of the folds held out as validation data. The final estimation was calculated as the averaged accuracy of the estimator for each fold. For this study 7-fold cross validation was used.

In cohort construction, features for epileptic patients were indexed by the date of first failure, and events (diagnoses, procedures, etc.) occurring previous to the index date were aggregated as features for ingestion by the unsupervised and supervised machine learning models. Patients with 4 or more future failures were considered as case, with control being epileptic patients with only one AED failure with no future failure. As the Events3 dataset was quite large, case and control patients were sampled from larger populations.

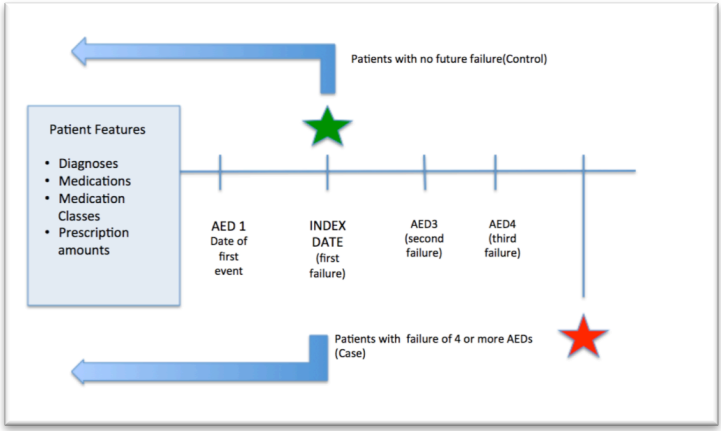

Figure 1: Cohort Construction

In sampling from the Events3 dataset, which was over 19gb in size, 58800 distinct refractory patients (case) were observed. From this population, 2000 case patients and 2000 control patients were subsampled to create balanced classes for the estimators. The raw features in the data contained over 20624 attributes, which when reduced using Kernel PCA, resulted in 6600 attributes of events.

In the case of feature selection, filtered events for case and control were transposed to create a row-wise feature matrix per patient. In this study little feature engineering was intentionally done as to evaluate the performance of unsupervised feature extraction:

1. $x = \begin{pmatrix} x_1 \\ \vdots \\ x_n \end{pmatrix}$

2. $X = \sum_{i=1}^{N} x^T$

**Approach and Implementation**

For this project, Spark 1.6 and Hadoop were configured on Google Cloud Dataproc, with Jupyter notebook integration. Data manipulation, aggregation, cleaning were completed using Spark SQL DataFrames, and machine learning models were developed and tested using scikit –learn in Python.

In setting up a Big Data environment, 1 High memory 32gb master node with 3 4gb worker nodes were provisioned. As this was a trial environment, there were limitations on the ability to scale out- or provision more worker nodes dynamically, which resulted in longer compute times. For all of the Spark SQL dataframe operations for example, the average compute time was around 6 hours on the event3 dataset.

**Experiment Design**

For the first hypothesis, unsupervised algorithms were run over the data to investigate accuracy in clustering techniques. In addition to running clustering algorithms over the raw features of the data, dimensionality reduction algorithms were used to project the data into lower dimensional spaces to investigate clustering efficacy.

The results did not show any significant accuracy of the clustering algorithms to find distinct cluster, both in the normal and reduced dimension spaces.

| Clustering Algorithm | Dimensionality Reduction Algorithm | Adjusted RAND Score | Adjusted Mutual Information Score |
|---|---|---|---|
| Kmeans | None | 0 | 0 |
| GMM | None | 0 | 0 |
| Spectral Clustering | None | -0.001448616 | -0.000387669 |
| Heirarchal Agglomerative Clustering | None | -0.000226729 | -8.20E-05 |
| Kmeans | PCA | 0 | 0 |
| GMM | PCA | 0 | 0 |
| Spectral Clustering | PCA | 0.000351541 | 0.000288785 |
| Heirarchal Agglomerative Clustering | PCA | -0.000226729 | -8.20E-05 |
| Kmeans | ICA | -0.000226729 | -8.20E-05 |
| GMM | ICA | -0.000226729 | -8.20E-05 |
| Spectral Clustering | ICA | -0.000783521 | -0.000799807 |
| Heirarchal Agglomerative Clustering | ICA | -0.000226729 | -8.20E-05 |
| Kmeans | KernelPCA | -0.001025027 | -0.000743991 |
| GMM | KernelPCA | -0.001228054 | -0.000864909 |
| Spectral Clustering | KernelPCA | -0.001093882 | -0.000786846 |
| Heirarchal Agglomerative Clustering | KernelPCA | -0.001094765 | -0.000764291 |
| Kmeans | ISOMAP | 0 | 0 |
| GMM | ISOMAP | 0 | 0 |
| Spectral Clustering | ISOMAP | -0.000997086 | -0.000287121 |
| Heirarchal Agglomerative Clustering | ISOMAP | -0.000226729 | -8.20E-05 |

Figure 2: Results from Clustering algorithms tested for hypothesis 1.

For hypothesis 2, a group of classification algorithms were tested to evaluate estimator performance on data reduced by Kernel PCA.

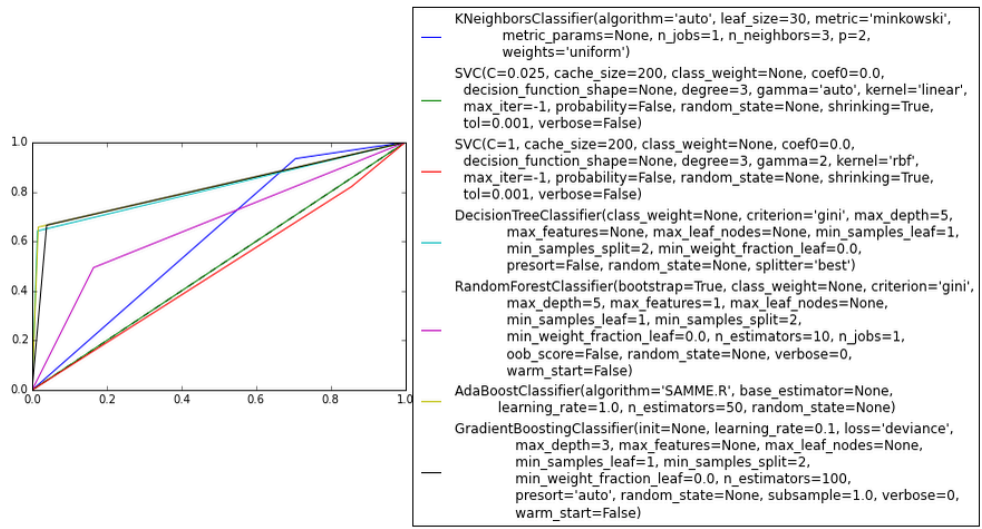

Figure 3: ROC Chart for Estimators trained on Kernel PCA reduced data.

In reviewing estimator performance over the reduced dataset, Tree-based classifiers, such as Decision Tree, ADABoost, and Ensemble Gradient Boosted Tress were observed to have the greatest AUC. An explanation for this could be that tree-based classifiers could capture the data that were non-linearly separable, thereby increasing performance. Where the decision boundary of the linear and kernel support vector machines and other distance-based classifiers did not capture the underlying non-linear quality in the data

**Kernel PCA**

Kernel PCA is a variation of standard Principal Components Analysis- where instead of extracting the eigenvectors of the largest eigenvalues of the covariance matrix:

$$3. \quad X' = \frac{1}{N}\sum_{i=1}^{N} x_i \, x_i T$$

The data is mapped into a higher dimensional space by a kernel function. This variation is also known as the "kernel trick":

$$4. \quad X' = \frac{1}{N}\sum_{i=1}^{N} \phi(x_i)\phi(x_i)T$$

In investigation for best hyperparameter selection: $\gamma$ is the parameter adjusted, given by:

$$5. \quad \phi(x_i, x_j) = \exp(-\gamma \, \|x_i - x_j\|_2^2)$$

In this Study, Gradient Boosted Decision Trees were observed to have significant classification accuracy.

**Gradient Boosted Decision Trees**

GBDT trains additive models of weak learners in a forward stage-wise fashion, where at each stage, weak learners are chosen to minimize a given loss function $\mathcal{L}$ given the current model and fit.

$$6. \quad F_m(X') = F_{m-1}(X') + \arg\min_h \sum_{i=1}^{n} \mathcal{L}(y_i, F_{m-1}(X'_i) - h(X'))$$

The negative gradient of the loss function is minimized until convergence, which is given by $\alpha$, where $\alpha$ is the hyperparameter to optimize:

$$7. \quad \alpha' = \arg\min_\alpha \sum_{i=1}^{n} \mathcal{L}(y_i, F_{m-1}(X'_i) - \alpha \frac{\partial \mathcal{L}(y_i, F_{m-1}(X'_i))}{\delta F_{m-1}(X'_i)})$$

In searching the hyperparameter space for the best setting for $\alpha$, a setting of 0.25 was found to have the best rate for estimator performance. Additional a tree depth of 5 was observed increase performance.

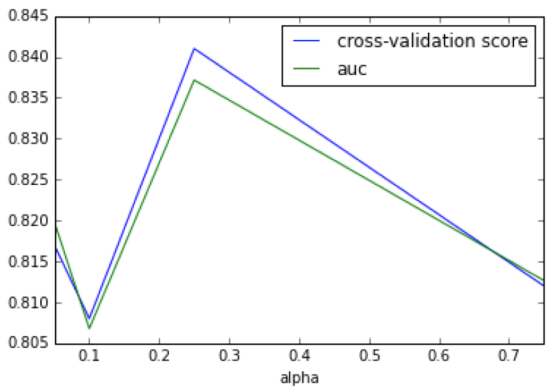

Figure 5: Classifier performance vs. changes in alpha

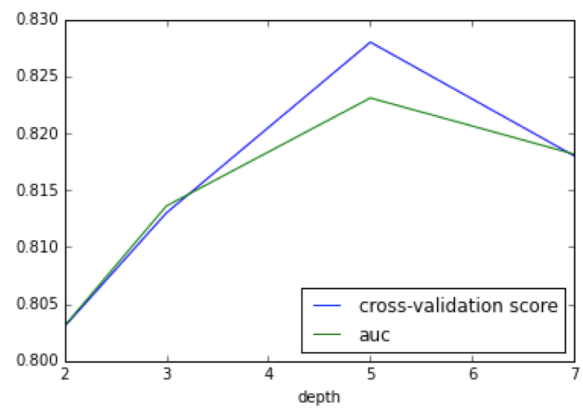

Figure 5: Classifier performance vs. changes in tree depth

An additional feature of using a Decision tree-type classifier for this problem is that the feature importance can be obtained and measured using the information gain of the features. In this study, the class of top features ranked on importance were previous diagnoses, indicating that previous diagnoses of comorbidity have a strong predictive quality to drug resistance. In this study, the feature names were disguised for security purposes, however it would be interesting to evaluate in future work whether or not these diagnoses are correlated with different modes of causality, such as the gene variant hypothesis [3], which indicates that P-glycoprotein, encoded by the ABCB1 gene, might play a role in AED resistance.

**Conclusion**

This study examined data for epileptic patients to refine methodologies for predicting drug resistance. A presentation of results is available at : https://youtu.be/3Cwj_EeMtLg

The challenges around this project were in dealing with large datasets in cloud environments, where the data were not easily modeled using linear classifiers.  In this case, it was discovered that transforming patient features using Kernel PCA and an Ensemble of Gradient Boosted Trees were able to capture some signal in the data. In reviewing our second hypothesis, we can conclude that we can create predictions around which patients may be more at risk of developing drug resistance with a reasonable degree of accuracy.

In reviewing hypothesis 1, clustering accuracy on raw features and features projected into lower dimensional spaces was not observed using the described methodologies. This section of the study did bring value, although we must reject the hypothesis under current methods attempted, as it was observed that using dimensionally reduced features improved classifier accuracy.

Further work could entail working with undisguised features to examine relationships in past diagnoses in relation to future drug resistance in support of the gene variant hypothesis [3].

**Supplemental Material**

- https://youtu.be/3Cwj_EeMtLg

**References**

1. Wiebe, S. & Jette, N. Pharmacoresistance and the role of surgery in difficult to treat epilepsy. *Nature Rev. Neurol.* **8**, 669–677 (2012).
2. Rogawski, M. A. The intrinsic severity hypothesis of pharmacoresistance to antiepileptic drugs. *Epilepsia* **54** (Suppl. 2), 32–39 (2013)
3. Löscher, W. Critical review of current animal models of seizures and epilepsy used in the discovery and development of new antiepileptic drugs. *Seizure* **20**, 359–368 (2011).
4. Pardoe HR, Cole JH, Thesen T, Blackmon K, Kuzniecky R. Abstract 1.146. Do seizures age the brain? Machine learning analysis of structural MRI. Presented at: American Epilepsy Society Annual Meeting; Dec. 4-8, 2015; Philadelphia.
5. Schmidt, D. & Löscher, W. Drug resistance in epilepsy: putative neurobiologic and clinical mechanisms. *Epilepsia* **46**, 858–877 (2005).
6. Hernández-Ronquillo L2, Buckley S1, Téllez-Zenteno JF. Predicting drug resistance in adult patients with generalized epilepsy: A case-control study. Epilepsy Behav. 2015 Dec;53:126-30. doi: 10.1016/j.yebeh.2015.09.027. Epub 2015 Nov 10.
7. Bernhard Schoelkopf, Alexander J. Smola, and Klaus-Robert Mueller. 1999. Kernel principal component analysis. In Advances in kernel methods, MIT Press, Cambridge, MA, USA 327-352.
8. http://stats.stackexchange.com/
9. http://pandas.pydata.org/pandas-docs
10. http://scikit-learn.org/stable/
11. Bernhard Schoelkopf, Alexander J. Smola, and Klaus-Robert Mueller. 1999. Kernel principal component analysis. In Advances in kernel methods, MIT Press, Cambridge, MA, USA 327-352.
12. Mitchell, Tom M. (1997). Machine Learning. The Mc-Graw-Hill Companies, Inc. ISBN 0070428077.